\documentclass[conference]{IEEEtran}
\IEEEoverridecommandlockouts

\usepackage{cite}
\usepackage[absolute,overlay]{textpos}
\usepackage{fancyhdr}
\usepackage{tikz}
\usepackage{tabularx}
\usepackage{booktabs}
\usetikzlibrary{positioning, arrows.meta}
\usepackage{amsmath,amssymb,amsfonts}
\usepackage{algorithmic}
\usepackage{graphicx}
\usepackage{textcomp}
\usepackage{xcolor}
\def\BibTeX{{\rm B\kern-.05em{\sc i\kern-.025em b}\kern-.08em
    T\kern-.1667em\lower.7ex\hbox{E}\kern-.125emX}}

\begin{document}

\title{Inference Time Feature Injection: A Lightweight Approach for Real-Time Recommendation Freshness}

\author{

\IEEEauthorblockN{1\textsuperscript{st} Qiang Chen}
\IEEEauthorblockA{\textit{Tubi}\\
San Francisco, United States \\
qiang@tubi.tv}

\and
\IEEEauthorblockN{2\textsuperscript{nd} Venkatesh Ganapati Hegde}
\IEEEauthorblockA{\textit{Tubi}\\
San Francisco, United States \\
vhegde@tubi.tv}

\and
\IEEEauthorblockN{3\textsuperscript{rd} Hongfei Li}
\IEEEauthorblockA{\textit{Tubi}\\
Beijing, China \\
hongfeili@tubi.tv}

}
\maketitle

\begin{textblock*}{\textwidth}(0cm,26cm) 
  \centering
  \footnotesize
  \copyright~2025 IEEE. Personal use of this material is permitted. Permission from IEEE must be obtained for all other uses, in any current or future media, including reprinting/republishing this material for advertising or promotional purposes, creating new collective works, for resale or redistribution to servers or lists, or reuse of any copyrighted component of this work in other works.
\end{textblock*}

\pagestyle{fancy}
\fancyhf{} 
\fancyfoot[C]{\footnotesize \copyright~2025 IEEE. Personal use of this material is permitted. 
Permission from IEEE must be obtained for all other uses, in any current or future media, 
including reprinting/republishing this material for advertising or promotional purposes, 
creating new collective works, for resale or redistribution to servers or lists, 
or reuse of any copyrighted component of this work in other works.}

\setlength{\footskip}{20pt}  

\begin{abstract}
Many recommender systems in long-form video streaming reply on batch-trained models and batch-updated features, where user features are updated daily and served statically throughout the day. While efficient, 
this approach fails to incorporate a user’s most recent actions, often resulting in stale recommendations. 
In this work, we present a lightweight, model-agnostic approach for intra-day personalization that selectively 
injects recent watch history at inference time without requiring model retraining.
Our approach selectively overrides stale user features at inference time using the recent watch 
history, allowing the system to adapt instantly to evolving preferences. By reducing the personalization 
feedback loop from daily to intra-day, we observed a statistically significant 0.47\% increase in key user engagement metrics which ranked among the most substantial engagement 
 gains observed in recent experimentation cycles. To our knowledge, this is the first published evidence that
  intra-day personalization can drive meaningful impact in long-form video streaming service, providing a compelling alternative 
  to full real-time architectures where model retraining is required.
  
\end{abstract}


\section{Introduction}
Personalization has become the backbone of content discovery in streaming platforms, where large libraries must be efficiently narrowed down to surface what a viewer is most likely to watch. In long-form video streaming, such as movies and TV shows, recommender systems play an especially high-stakes role because poor recommendations not only risk session abandonment, but also undermine monetization for ad-supported services.

A key axis of optimization in recommender systems is freshness: how quickly the system adapts to reflect a user’s evolving preferences. Most large-scale platforms use batch-trained models, which update user features once per day or less frequently. While batch systems offer strong stability and are easy to scale, they introduce a fundamental latency between user behavior and the recommendations the user receives. A user might finish a thriller in the morning but still see comedy suggestions from the previous evening’s binge. This lag in personalization creates friction in the viewing experience.

In response, real-time recommender systems have emerged in domains where recency is paramount-such as e-commerce, advertising, and video streaming \cite{aws2023, doordash2023, DNYoutubeRec}. These systems leverage streaming infrastructure and online feature stores to capture and respond to user signals within seconds or minutes. In these high-frequency environments, fresh personalization has been shown to improve engagement and relevance. However, real-time systems also impose significant engineering overhead, making them complex to deploy and maintain.

Long-form video streaming presents a more ambiguous tradeoff. User feedback is sparse, sessions are long, and preferences may not shift as quickly as in short-form media. Although companies like Netflix and Amazon have developed real-time or nearline components to support session-level adaptation \cite{netflix2025Rapid}, no prior work has empirically validated whether intra-day freshness. i.e., personalization within hours rather than a full day-actually drives meaningful engagement gains in long-form contexts.

This paper addresses that gap. We introduce a lightweight, model-agnostic approach that selectively overrides stale user watch history features with fresh signals at inference time, without retraining the model. Our method reduces the feedback loop from a day to intra-day, yielding a statistically significant 0.47\% increase in key user engagement metrics on a major production ads supported video on demand (AVOD) platform.

To our knowledge, this is the first published evidence that intra-day personalization can move the needle in long-form streaming. Our findings offer a pragmatic alternative to full real-time architectures where model retraining is required, demonstrating that even modest reductions in personalization latency can unlock measurable engagement gains at scale.
\section{Related Works}

\subsection{Personalization in Long-Form Streaming}
Most recommender systems in long-form video rely on batch inference and daily model refreshes \cite{estuary2023, spotintelligence2025}. This architecture is designed for stability and throughput, aggregating user behavior logs and retraining model components at fixed cadences. It is well-suited for domains where user preferences evolve slowly and feedback is sparse. However, the lag between action and adaptation introduces friction in a setting where viewer intent can change dramatically even within a single session.
While companies like Netflix have discussed hybrid personalization stacks that incorporate session-aware re-ranking or post-processing layers \cite{netflix2025Rapid}, the underlying ranking logic remains driven by batch-trained models. More importantly, there is little published work assessing whether increased freshness specifically in the form of intra-day personalization leads to measurable engagement improvements in long-form streaming.

\subsection{Real-Time Personalization in Adjacent Domains}
In contrast, personalization in high-frequency interaction domains has embraced real-time modeling. Ad-serving systems use streaming features to adjust ranking based on impression-level feedback \cite{aws2023}. Short-form video platforms like YouTube Shorts and Kuaishou integrate behavioral signals such as dwell time, swipes, and skips at second-level resolution to tailor the next recommendation \cite{DNYoutubeRec}. E-commerce platforms apply similar techniques to personalize in-session browsing \cite{doordash2023}.
These real-time architectures typically depend on online feature stores and streaming ingestion frameworks \cite{feast2023, yu2019adaptive}, allowing fresh signals to be joined with static features during inference. Despite the success of this pattern in short-form and transactional settings, there is no published evidence showing that real-time signal integration improves engagement in long-form video platforms.

\subsection{Bridging the Latency Gap: Intra-Day Personalization}
This paper addresses the open question of whether freshness matters in long-form personalization, and if so, how much infrastructure is truly needed to deliver it. We present a technique that selectively injects fresh session-level signals into the ranking pipeline at inference time, bypassing the need for model retraining.
Our approach reduces personalization latency from daily cycles to intra-day intervals using a lightweight feature injection mechanism. In a large-scale A/B test on a production AVOD platform, this technique yielded a statistically significant lift in key user engagement metrics. This result demonstrates that intra-day personalization can deliver meaningful value.

\subsection{Practical Tradeoffs in Long-Form Personalization}
Real-time systems offer best-in-class responsiveness, but they introduce substantial operational and engineering overhead. For many long-form platforms, these costs may outweigh the perceived benefits. A modest reduction in latency achieved through simple inference-time injection can yield high return on investment, particularly in long-form settings where real-time infrastructure is uncommon and intra-day freshness alone can deliver measurable engagement gains

\section{Proposed Method: Inference Time Feature Injection}

Modern recommender systems typically follow a two-stage pipeline\cite{borisyuk2016casmos, YouTubeVideoRec}: candidate retrieval followed by item ranking. Both stages rely on user-level features derived from user behaviors, such as watch history. For simplicity and efficiency, many large-scale systems employ periodic offline batch pipelines to process user behavior logs into usable features. These batch processes aggregate user interactions and generate features on a fixed schedule--often daily--which are then consumed by downstream recallers and ranking models.

While effective for scaling and model stability, this architecture imposes a critical limitation: recommendations are only as fresh as the most recent batch. When a user exhibits new behavior after the last batch run, the system cannot adapt until the next scheduled processing cycle. This latency results in stale recommendations, particularly in scenarios where user intent evolves rapidly.

\subsection{Baseline: Batch-Based Recommendation}

\begin{figure}[h]	
	\centering	
	\includegraphics[width=0.5\textwidth]{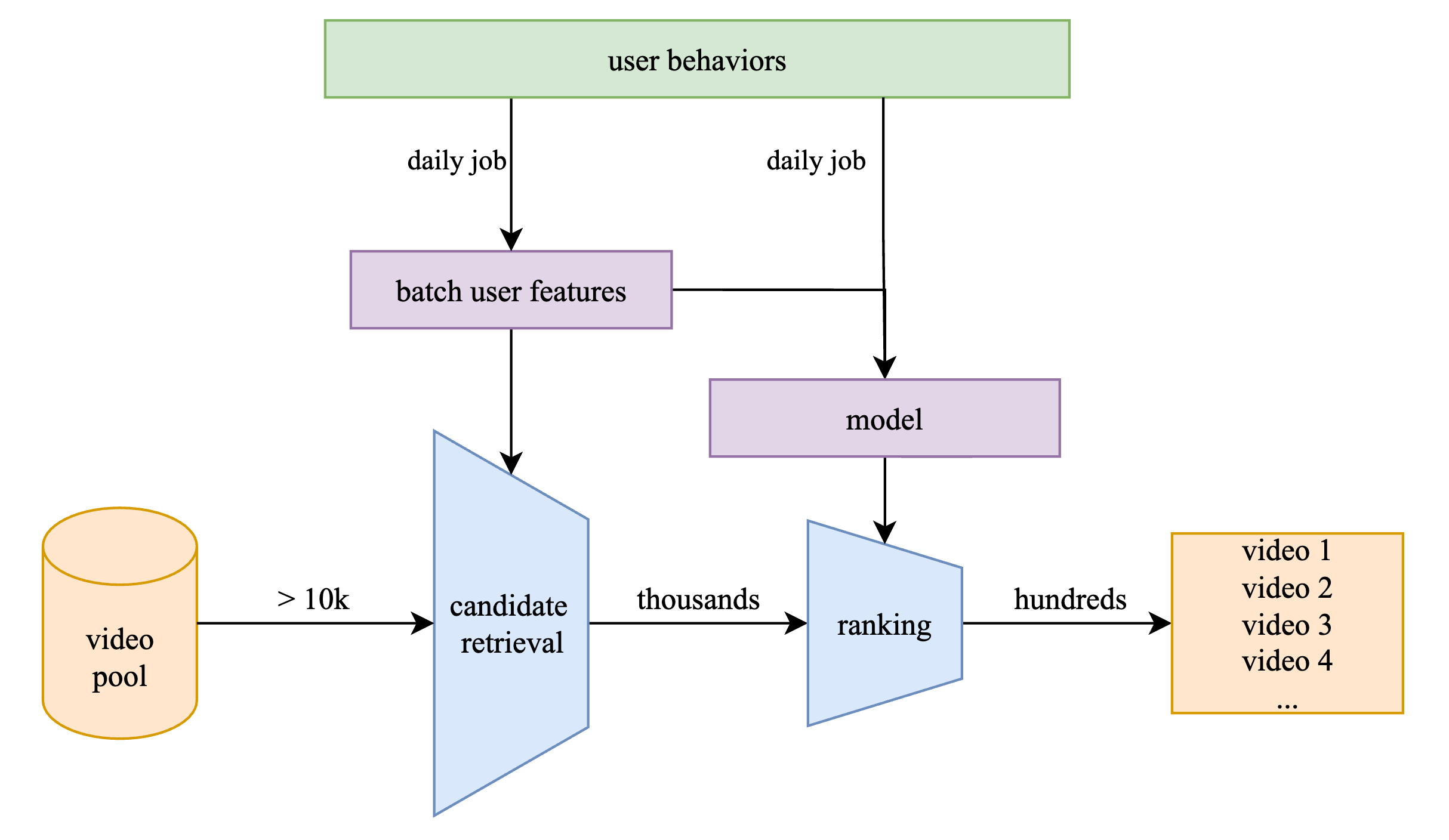}
	\caption{Two-stage recommendation architecture with batch-trained models and batch-updated features. Daily jobs process user behavior and then generate features consumed by downstream recallers and ranking models.}
	\label{fig:ranking_process_control}
\end{figure}

The standard batch-based recommendation system operates as shown in Figure \ref{fig:ranking_process_control}.

\begin{enumerate}
    \item \textbf{Candidate Retrieval Stage:} The primary recaller uses the user's watch history -- updated through daily batch processing -- to retrieve a set of similar or relevant items. Additional recallers (e.g., popularity-based ) are used to diversify the candidate pool.
    
    \item \textbf{Ranking Stage:} For each candidate item, features are constructed using the batch-generated user history, item metadata, and contextual information. These features are then passed to a pre-trained ranking model to produce a final recommendation list.
\end{enumerate}

This system ensures high scalability and reproducibility but suffers from delayed responsiveness. User interactions that occur after the most recent batch are effectively invisible to the model until the next processing cycle.

\subsection{Proposed Enhancement: Feature Injection during Inference}

\begin{figure}[h]
	\centering	
	\includegraphics[width=0.5\textwidth]{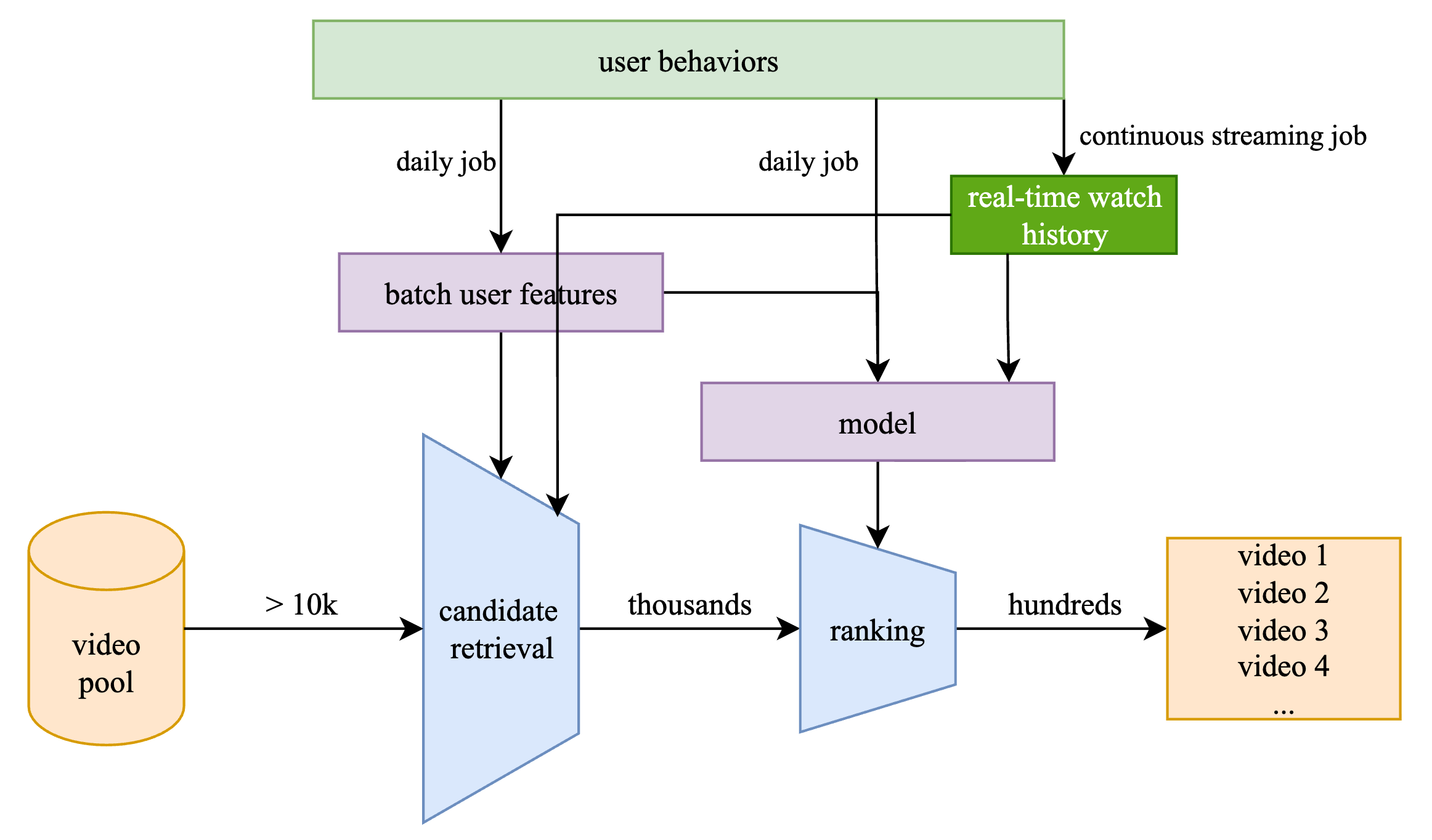}
	\caption{Two-stage recommendation architecture with inference time user watch history injection. A dedicated real-time feature service was implemented, it is a continuous streaming job that continuously consumes
 user behavior events and transforms them into model-ready real-time watch history features with minimal delay.}		\label{fig:ranking_process_treatment}
\end{figure}

To improve responsiveness without sacrificing model stability, we introduce a minimal architectural extension that enables real-time personalization through inference time recent user watch history injection. This approach merges user’s batch-updated watch history and the recent watch history, and then inject them as if it is batch-updated watch history, while preserving the existing batch-trained model as shown in Figure \ref{fig:ranking_process_treatment}.

\begin{enumerate}
    \item \textbf{Candidate Retrieval Stage:} The main recaller is enhanced to incorporate the user's recent watch
	 history alongside their existing batch-updated watch history. This allows the system to retrieve items that are more aligned with the user's immediate intent. Auxiliary recallers remain unchanged. 
    
    \item \textbf{Ranking Stage:} During inference, we dynamically merge two types of user history: (a) the batch-updated user watch history and (b) the recent user watch history in real-time. These are combined, which is then passed into the existing ranking model. No retraining is required.
\end{enumerate}

A dedicated real-time feature service was implemented to support this architecture. The service continuously consumes
 user interaction events and transforms them into model-ready features with minimal delay.

The daily batch history process can handle a large scale of data and including long period of them but their 
responsiveness is low, the real-time feature service on the other hand can only maintain a short time range, 
but they can response to latest user behavior within seconds. So we combine two of them to achieve both purpose,
 long time range and responsiveness to latest behavior. 

\subsection{Design Rationale and Trade-Offs}

Traditionally, maintaining strict feature consistency between training and inference is considered best practice.
 However, we argue that a controlled deviation--where only inference time features are updated in real time--can 
 offer practical advantages:

\begin{itemize}
    \item \textbf{Engineering Simplicity:} The training pipeline remains unchanged, avoiding the complexity of 
	adding streaming features into model retraining workflows\cite{zhu2024interest}. Some other approaches \cite{liu2022monolith} 
	even require online training, which requires more resources and is more complex.
    \item \textbf{Prediction Relevance:} Injecting fresh behavior at inference directly addresses the temporal 
	mismatch between stale batch-updated user features and recent user interest, improving personalization in the moments that matter most.
    \item \textbf{Interpretability:} This method acts as a “temporal acceleration” mechanism--delivering recommendations reflective of user behavior that would traditionally only be incorporated in the next-day batch. In effect, it brings tomorrow’s recommendations into today’s experience.

\end{itemize}

\section{Result and Discussion}

An online A/B experiment was conducted on Tubi's production large scale recommendation system to evaluate the effectiveness of the proposed inference time feature injection technique.
In the control group, recommendations were generated using batch-updated user watch history with a typical refresh 
latency of 24 hours. In contrast, the treatment group received recommendations based on real-time watch history, 
injected directly at inference time. This modification enabled the system to incorporate the latest session-level signals without retraining the underlying model.

The treatment group exhibited a statistically significant 0.47\% lift in key user engagement metrics which ranked among the most impactful changes deployed within recent experimentation cycles underscoring 
 the value of improved responsiveness in a long-form content setting.

While exploring the inference time history injection, another approch was tested, which has elegant online and offline 
feature consistency, in this variant, we added auxiliary features explicitly representing recent watch behavior
 (e.g., items watched in the past few hours), ensuring that these features were included both during model training 
 and inference. Despite maintaining semantic consistency across stages, this method did not yield measurable gains 
 in engagement metrics. We hypothesize that model training tend to fit previuos model recommendation instead of learning what user really like. This outcome suggests the effectiveness of inference time history injection method.
 
A key insight from our experiments is that strict alignment between training and inference features--long assumed 
to be best practice--may not always yield the best outcomes. Our approach, which introduces a controlled feature 
discrepancy by injecting real-time signals only at inference, proved both simpler to implement and more effective 
in practice. This design allows us to retain the stability of batch-trained models while achieving measurable responsiveness
 at minimal cost.

Additionally, the system is robust to sparsity and does not require significant reengineering of existing model
 architecture or retraining logic. By keeping the real-time signals limited to recent watch history and merging 
 them dynamically with batch features, we avoid introducing instability or noise into the model while enhancing 
 its temporal relevance.

These findings highlight a critical trade-off in long-form recommender systems: the need to balance personalization
 freshness with model stability. Our work shows that inference time feature injection offers a practical and lightweight solution to this challenge.

\bibliographystyle{IEEEtran}
\bibliography{myref}
\end{document}